\documentclass[runningheads]{llncs}
\usepackage{cite}
\usepackage{amsmath,amssymb,amsfonts}
\usepackage{graphicx}
\usepackage{textcomp}
\usepackage{xcolor}
\usepackage{bm}
\usepackage{tabularx}
\usepackage{multirow}
\usepackage{tikz}
\usetikzlibrary{positioning, calc, fit, shapes, arrows, external}

\usepackage[caption=false]{subfig}
\usepackage [english]{babel}
\usepackage [autostyle, english = american]{csquotes}
\MakeOuterQuote{"}

\begin{document}
\title{Improving Prediction Confidence in Learning-Enabled Autonomous Systems\thanks{This  work  is  supported  in  part  by  AFOSR  DDDAS  through contract FA9550-18-1-0126 program and DARPA through contract number FA8750-18-C-0089. Any opinions, findings, and conclusions or recommendations expressed are those of the author(s) and do not necessarily reflect the views of the sponsor.}}
\author{Dimitrios Boursinos \and
Xenofon Koutsoukos}
\authorrunning{D. Boursinos and X. Koutsoukos}
\institute{Vanderbilt University, Nashville TN, USA\\
\email{\{dimitrios.boursinos,xenofon.koutsoukos\}@vanderbilt.edu}
}
\maketitle              % typeset the header of the contribution

\begin{abstract}
Autonomous systems use extensively learning-enabled components such as deep neural networks (DNNs) for prediction and decision making. 
%However, modern DNN architectures are very complex and reasoning about the their behavior is very challenging. 
In this paper, we utilize a feedback loop between learning-enabled components used for classification and the sensors of an autonomous system in order to improve the confidence of the predictions. We design a classifier using Inductive Conformal Prediction (ICP) based on a triplet network architecture in order to learn representations that can be used to quantify the similarity between test and training examples. The method allows computing confident set predictions with an error rate predefined using a selected significance level. A feedback loop that queries the sensors for a new input is used to further refine the predictions and increase the classification accuracy. The method is computationally efficient, scalable to high-dimensional inputs, and can be executed in a feedback loop with the system in real-time. The approach is evaluated using a traffic sign recognition dataset
and the results show that the error rate is reduced. 

\end{abstract}

\section{Introduction}
\label{sec:introduction}
%motivation
Autonomous systems are equipped with sensors to observe the environment and take control decisions. Such systems can benefit from methods that allow to improve prediction and decision making through a feedback loop that queries the sensor inputs when more information is needed~\cite{1398021}. Such a paradigm has been used in a variety of applications such as multimedia context assessment~\cite{Aved2015}, aerial vehicle tracking~\cite{7471414}, automatic target recognition~\cite{10.1117/12.2016338}, self-aware aerospace vehicles~\cite{ALLAIRE20131959},  and smart cities~\cite{fujimoto2016dynamic}. 
%
% Machine Learning components are being used extensively in autonomous systems because of their ability to handle dynamic and uncertain environments. Modern Deep Neural Networks are parameterized with millions of values and are able to approximate complex functions allowing them to be used in applications with very high dimensional inputs like images and videos. This makes them a very good fit as perception components for self-driving vehicles equipped with multiple cameras and LiDARs to be able to recognize areas of interest like other vehicles, persons and traffic signs. These advances come with challenges regarding safety, as these models are not transparent and hard to understand how decisions are made.   
%
%problem
In particular, autonomous systems can utilize perception learning-enabled components (LECs) to observe the environment and make predictions used for decision making and control. LECs such as deep neural networks (DNNs) can generalize well on test data that come from the same distribution as the training data and their predictions can be trusted. However, during the system operation the input data may be different than the training data resulting to large prediction errors. An approach to address this challenge is to quantify the uncertainty of the prediction and query the sensors for additional inputs in order to improve the confidence of the prediction. The approach must be computationally efficient so it can be executed in real-time for closing the loop with the system.

%related work
Computing a confidence measure along with the model's predictions is essential in safety critical applications where we need to take into account the cost of errors and decide about the acceptable error rate. Neural networks for classification typically have a softmax layer to produce probability-like outputs. However, these probabilities cannot  be used reliably 
%as the true probabilities that a class will occur 
as they tend to be too high, they are overconfident, even for inputs coming from the training distribution~\cite{Guo:2017:CMN:3305381.3305518}. 
%Because of this, the softmax probabilities are not well-calibrated meaning they do not provide a good estimate of the error rates. 
The softmax probabilities can be calibrated to be closer to the actual probabilities scaling them with factors computed from the training data. Different methods that have been proposed to compute scaling factors include temperature scaling~\cite{Guo:2017:CMN:3305381.3305518}, Platt scaling \cite{Platt99probabilisticoutputs}, and isotonic regression \cite{Zadrozny:2002:TCS:775047.775151}. Although such methods can compute well-calibrated confidence values, it is not clear how they can be used for querying the sensors for additional inputs.
%that will limit the times that a decision cannot be made. 
Conformal prediction (CP) is another framework used to compute set predictions with well-calibrated error bounds~\cite{balasubramanian2014conformal}. The set predictions can be computed efficiently 
% using the inductive conformal prediction (ICP) that 
leveraging a calibration data set~\cite{papadopoulos2008inductive}.  
%that is used to compute the confidence values of new previously unseen inputs efficiently
However, such approaches do not scale for high-dimensional inputs such as camera images. 
In our prior work, we have developed methods handling high-dimensional inputs using inductive conformal prediction (ICP)~\cite{boursinos2020assurance,boursinos2020trusted}.

%contribution
This paper extends our prior work 
by designing a feedback loop between LECs used for classification and the sensors of an autonomous system in order to improve the confidence of the predictions. We design a classifier using ICP based on a triplet network architecture in order to learn representations that can be used to quantify the similarity between test and training examples. Given a significance level, the method allows computing confident set predictions. A feedback loop that queries the sensors for a new input is used to further refine the predictions and increase the classification accuracy. The method is computationally efficient, scalable to high-dimensional inputs, and can be executed in a feedback loop with the system in real-time. The approach is evaluated using a traffic sign recognition dataset and the results show that the error rate is reduced. 

%The main contribution of this paper is the use of DDDAS concepts along with the ICP framework. A feedback loop is added between the ICP classifier decision-making and the input sensors that allow the classifier to query for a new inputs when a confident classification cannot be made based on the current and previous inputs. Care was taken so that the inputs can be high-dimensional and the classifications can be made in real-time. We investigate the relationship between the classification error-rate, the classification time and the feedback-loop parameters. Finally, we evaluate the feedback-loop ICP classifier applying it to an autonomous vehicle to recognize traffic signs in real-time.
\section{Triplet-based ICP}
\label{sec:icp}
\begin{figure}[htb]
\centering
\tikzstyle{block} = [draw, fill=orange!60, rectangle, 
    minimum height=3em, minimum width=3em]
\begin{tikzpicture}[auto, node distance=2cm,>=latex']
    \node [block, fill=green!40] (env) {Environment};
    \node [block, right of=env, node distance=3.4cm, align=center] (classification) {Triplet based\\ICP};
    \node [block, right of=classification, node distance=3.7cm, align=center] (decision_making) {Decision\\Making};
    \node [block, fill=blue!30, right of=decision_making, node distance=3.3cm, align=center] (controller) {Autonomous\\System};
    
    \draw[->] (env) -- (classification) node[midway,sloped,below,align=center] {sensor\\inputs};
    \draw[->] (classification) -- (decision_making) node[midway,sloped,below,align=center] {prediction\\sets};
    \draw[->] (decision_making) -- (controller) node[midway,sloped,below,align=center] {decision};
    \draw[->] ($(decision_making.south east)!0.50!(decision_making.south west)$) -- +(0,-1) -- ($(env.south east)!0.50!(env.south west)+(0,-1)$) node[midway,sloped,below,align=center] {query\\sensors} -- ($(env.south east)!0.50!(env.south west)$);
\end{tikzpicture}
\caption{Feedback loop between the decision-making process and sensing}
\label{fig:approach_diagram}
\end{figure}
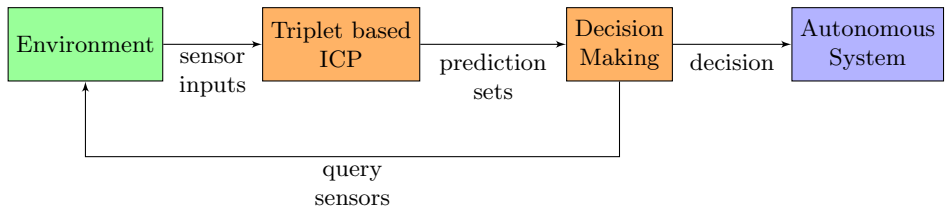

We consider an autonomous system that takes actions based on its state in the environment
as shown in Figure~\ref{fig:approach_diagram}. For example, a self-driving vehicle needs to take control actions based on the traffic signs it encounters. We design a classifier using ICP based on a triplet network architecture in order to learn representations that can be used to quantify the similarity between test and training examples. Given a significance level, the method allows computing confident set predictions. A feedback loop that queries the sensors for a new input is used to further refine the predictions and increase the classification accuracy.

Triplet networks are DNN architectures trained to learn representations of the input data for distance learning~\cite{hoffer2014deep}. The last layer of a triplet network computes a representation $Net(x)$ of the input $x$. For training, a triplet network is composed using three copies of the same neural network with shared parameters. It is trained on batches formed with triplets of data points. Each of these triplets has an anchor data point $x$, a positive data point $x^+$ that belong to the same class as $x$ and a negative data point $x^-$ of a different class. The objective is to maximize the distance between inputs of different classes $|Net(x)-Net(x^-)|$ and minimize the distance of inputs belonging to the same class $|Net(x)-Net(x^+)|$. To achieve this, training uses the loss function: 
\begin{equation*}
    Loss(x,x^+,x^-)=max(|Net(x)-Net(x^+)|-\\|Net(x)-Net(x^-)|+\alpha,0)
\end{equation*}
where $\alpha$ is the margin between positive and negative pairs.

The simplest way to form triplets is to randomly sample anchor data points from the training set and augment them by randomly selecting one training sample with the same label as the anchor and one sample with a different label. However, for many of these ($x, x^+, x^-$) triplets $|Net(x)-Net(x^-)|>>|Net(x)-Net(x^+)|+\alpha$, which provides very little information for distance learning and leads to slow training and poor performance. The training can be improved by carefully mining the training data \cite{xuan2019improved}. For each training iteration, first, the anchor training samples are randomly selected. For each anchor, the hardest positive sample is chosen, that is a sample from the same class as the anchor that is located the furthest away from the anchor. Then, the triplets are formed by mining all the hard negative samples, that is the samples that satisfy $|Net(x)-Net(x^-)|<|Net(x)-Net(x^+)|$. When the training is completed, only one of the three identical DNN copies is used to map an input $x$ to its embedding representation $Net(x)$.

Consider a training set \{$z_1,\dots,z_l$\}, where each $z_i\in Z$ 
is a pair $(x_i,y_i)$ with $x_i$ the feature vector and $y_i$ the label. We also consider a test input $x_{l+1}$ which we wish to classify. 
%The task is to make classifications that will satisfy a chosen significance level. 
The underlying assumption of ICP is that all examples ($x_i,y_i$), $i=1,2,\dots$ are independent and identically distributed (IID) generated from the same but typically unknown probability distribution.
For a chosen classification significance level $\epsilon\in [0,1]$, ICP generates a set of possible labels $\Gamma^\epsilon$ for the input $x_{l+1}$ 
%with ground truth label $y_{l+1}$, 
such that $P(y_{l+1}\notin\Gamma^\epsilon)<\epsilon$. 

Central to the framework is the use of \textit{nonconformity measures} (NCM), a metric that indicates how different an example $z_{l+1}$ is from the examples of the training set $z_1,\dots,z_l$. A NCM that can be computed efficiently in real-time is the \textit{k-Nearest Neighbors} ($k$-NN)~\cite{papernot2018deep} defined in the embedding space generated by the triplet network.
%and can be used efficiently in the ICP implementation .
%, (2) the \textit{one Nearest Neighbor} (1-NN)~\cite{Vovk:2005:ALR:1062391} and (3) the \textit{Nearest Centroid}~\cite{balasubramanian2014conformal}.
The $k$-NN NCM finds the $k$ most similar examples in the training data and counts how many of those are labeled different than the candidate label $y$ of a test input $x$. We denote $f:~X\rightarrow~V$ the mapping from the input space $X$ to the embedding space $V$ defined by the triplet's last layer. After the training of the triplet is complete, we compute and store the encodings $v_i=f(x_i)$ for the training data $ x_i $. Given a test example $x$ with encoding $v=f(x)$, we compute the $k$-nearest neighbors in $V$ and store their labels in a multi-set $\Omega$. The $k$-NN nonconformity of the test example $ x $ with a candidate label $y$ is defined as:
$$\alpha(x,y)=|i\in\Omega:i\neq y|$$

% The 1-NN NCM requires to find the most similar example in the training set that has the same label as the candidate label $y$ of a test input $x$ as well as the most similar example that belong to any other class other than $y$. It is defined as: 
% $$\alpha(x,y)=\dfrac{\min_{i=1,\dots,n:y_i=y}d(v,v_i)}{\min_{i=1,\dots,n:y_i\neq y}d(v,v_i)}$$
% where $ v = f(x) $, $v_i=f(x_i)$, and $d$ is the euclidean distance metric in the $V$ space.

% The Nearest Centroid NCM simplifies the task of computing individual training examples that are similar to a test example when there is a large amount of training data. We expect examples that belong to a particular class to be similar to each other so for each class $y_i$ we compute its centroid $\mu_{y_i}=\dfrac{\sum_{j=1}^{n_i}v_j^i}{n_i}$, where $v_j^i$ is the embedding representation of the  $ j^{th} $ training example from class $y_i$ and $n_i$ is the number of training examples in class $y_i$. The nonconformity function is defined as:
% $$\alpha(x,y)=\dfrac{d(\mu_y,v)}{\min_{i=1,\dots,n:y_i\neq y}d(\mu_{y_i},v)}$$
% where $ v = f(x) $. It should be noted that for computing the nearest centroid NCM, we need to store only the centroid for each class.

For statistical significance testing, $p$-values are assigned based on the computed NCM scores using a calibration set of labeled data that are not used for training.
% Larger NCM values $a(x_{l+1},y_{l+1})$ indicate greater dissimilarity between a test input $x_{l+1}$ with possible label $y_{l+1}$ and the training data $z_1\dots z_l$. To make sense of this values and assign p-values to each candidate label that correspond to confidence, we compare it with the NCM values computed using a \textit{calibration set} of known labeled data. 
The training set ($z_1\dots z_l$) is split into two parts, the \textit{proper training} set ($z_1\dots z_m$) of size $m<l$ that is used for the training of the triplet network and the \textit{calibration set} ($z_{m+1}\dots z_l$) of size $l-m$ that is used only for the computation of the $p$-values. The empirical $p$-value assigned to a possible label $j$ of an input $x$ is defined as the fraction of nonconformity scores of the calibration data that are equal or larger than the nonconformity score of a test input:
$$p_j(x)=\dfrac{|\{\alpha\in A:\alpha\geq\alpha(x,j)\}|}{|A|}.$$
The $p$-values are used to form the sets of candidate labels for a given significance level $\epsilon$. The label $j$ is added to $\Gamma^\epsilon$ if $ p_j(x) > \epsilon $.

\section{Feedback-loop for Querying the Sensors}
\label{sec:feedback}
Only the prediction sets $\Gamma^\epsilon$ that have exactly one candidate label can directly be used towards the final decision. When $|\Gamma^\epsilon|\neq1$ the approach queries the sensors for a new input. Incorrect classifications are more likely to happen during the first time steps of the process as every sensor input offers new information that may lead to a more confident prediction. For example, in the traffic sign recognition task, it is more likely for an incorrect classification to happen when the sign is far away from the vehicle and the image has low resolution as shown in Figure~\ref{fig:sign_time}. To avoid such incorrect classifications, in our method the final decision is made only after $k$ consecutive identical predictions. The parameter $k$ represents a trade-off between robustness and decision time, as larger $k$ leads to additional delay but more confident decisions. Further, very low $k$ values may lead to incorrect decisions while very large values may not allow a timely a decision.

The ICP framework produces well-calibrated prediction sets $\Gamma^\epsilon$ when inputs are IID. Depending on how small the chosen significance level is, $\Gamma^\epsilon$ may include a different number of candidate labels. The classification of an input requires $|\Gamma^\epsilon|=1$. In our previous work \cite{boursinos2020assurance,boursinos2020trusted}, we use a labeled validation set to compute the minimum significance level $\epsilon$ to reduce the prediction sets with more than one candidate label. However, in dynamic systems, sensor measurements change over time. Each new input in a sequence is related to previous inputs and the inputs are not IID. In this case, even though the calculated significance level $\epsilon$ will not lead to $|\Gamma^\epsilon|>1$, the actual error rate may not be bounded by $\epsilon$.

The main idea is to utilize a feedback-loop in order to lower the error-rate. %The quality of the data we wish to classify can variate depending on different things like sensor noise and the current conditions in the environment where the system operates. 
In order to reduce the incorrect predictions that may occur especially for low quality inputs, we require that $|\Gamma^\epsilon|=1$ with identical single candidate label for $ k $ consecutive sensor measurements. When this condition is satisfied for an input sequence, the prediction can be used for decision making by the autonomous system.
\section{Evaluation}
\label{sec:evaluation}
% In this section, we evaluate the challenges of applying ICP to sequential inputs and how the introduction of feedback loops lead to improved accuracy in sequential image recognition application for autonomous vehicles. Autonomous vehicles use a number of different sensors to observe the environment surrounding them and take the appropriate actions. A very important task, essential for safe operation is the recognition of traffic signs with a bounded error rate. We evaluate the challenges of applying ICP to sequential inputs and howcompare how the different nonconformity functions perform in terms of error rate per significance level, computational requirements and execution times and experiment with different DDDAS designs. 

% \vspace{0.1in}

\begin{figure}[tb]
\centering
\input{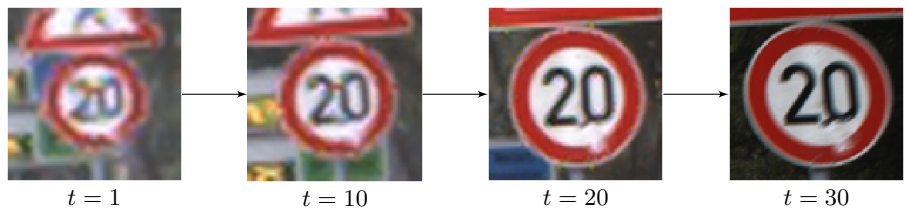}
\caption{Traffic sign over time (in frames)}
\label{fig:sign_time}
\end{figure}

\noindent
\textbf{Experimental Setup} We apply the proposed method to the German Traffic Sign Recognition Benchmark (GTSRB). A vehicle uses an RGB camera to recognize the traffic signs that are present in its surroundings. The dataset consists of 43 classes of signs and provides videos with 30 frames as well as individual images. The data are collected in various light conditions and include different artifacts like motion blur. The image resolution depends on how far the sign is from the vehicle as shown in Figure~\ref{fig:sign_time}. Since the input size depends on the distance between the vehicle and the sign, we convert all inputs to size 96x96x3. 10\% of the available sequences is randomly sampled to form the sequence test set. 10\% of the individual frames is randomly sampled to form another test set. All the remaining frames are shuffled and 80\% of them are used for training and 20\% are used for calibration and validation. 

The triplet network is formed using three identical convolutional DNNs with shared parameters. We use a modified version of the VGG-16 architecture using only the first four blocks because of the reduced input size. A dense layer of 128 units is used to generate the embedding representation of the inputs. All the experiments run in a desktop computer equipped with and Intel(R) Core(TM) i9-9900K CPU and 32 GB RAM and a Geforce RTX 2080 GPU with 8 GB memory.

%\vspace{0.05in}

\begin{table}[ht]
\centering
\begin{tabular}{|c|c|c|c|}
\hline
Training Accuracy & Validation Accuracy & IID Testing & Sequence Testing  \\ \hline
0.991             & 0.987               & 0.986                & 0.948                     \\ \hline
\end{tabular}
\caption{Triplet-based classifier performance}
\label{tab:model_accuracy}
\end{table}
\vspace{-7mm}

\noindent
\textbf{Model Performance} The triplet network can be used for classification of inputs using a $k$-Nearest Neighbors classifier in the embedding space. We first investigate how well the triplet network classifier is trained looking at the accuracy of the two test sets. One basic hypothesis of machine learning models is that the training and testing data sets should consist of IID samples. This is confirmed in Table \ref{tab:model_accuracy} where the accuracy for the testing set of IID examples is similar to the training accuracy while the testing accuracy for the set that includes sequences is lower.

\begin{figure}[ht]
\centering
\includegraphics[width=0.9\linewidth]{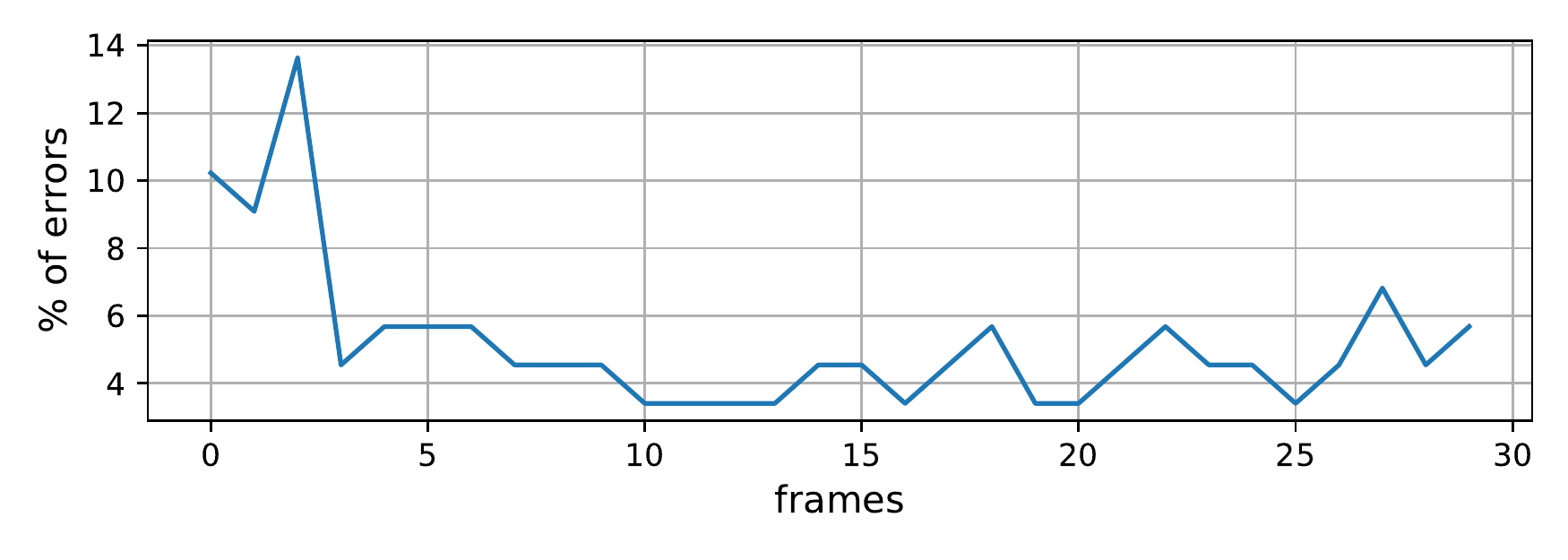}
\caption{Average error per frame for all the test sequences}
\label{fig:sequence_error}
\end{figure}
%\vspace{-5mm}

In order to investigate which frames are responsible for the larger error-rate in the sequences we plot the average error-rate per frame for the 30 frames of all the test sequences in Figure~\ref{fig:sequence_error}. The early frames of each sequence tend to have more incorrect classifications as expected since the sign images have lower resolution. 

\begin{table}[ht]
\centering
\begin{tabular}{c|c|c||c|c|}
\cline{2-5}
\multicolumn{1}{l|}{}                         & \multicolumn{2}{c||}{IID Test} & \multicolumn{2}{c|}{Sequences Test} \\ \hline
\multicolumn{1}{|c||}{$\epsilon$} & Errors       & Multiples      & Errors          & Multiples         \\ \hline
\multicolumn{1}{|c||}{0.017}                   & 1.7\%        & 0\%            & 5.6\%           & 0\%               \\ \hline
\end{tabular}
\caption{Triplet-based ICP performance for individual frames}
\label{tab:icp_performance}
\end{table}
\vspace{-7mm}

\noindent
\textbf{ICP Performance}
We apply ICP on single inputs to understand how the classifier performs without the feedback loop. The ICP is evaluated for both test sets in Table~\ref{tab:icp_performance}. An error corresponds to the case when the ground truth for a sensor input is not in the computed prediction set. We compute the smallest significance level $\epsilon$ that does not produce sets of multiple classes using the validation set.  Similar to the point classifier, the ICP classifier produces well-calibrated predictions only for the IID test inputs.

%the cl The triplet network can be used as a classifier when combined with a k-NN classifier. As the vehicle approaches a traffic sign, the ICP classifier makes predictions that are independent from previous predictions. The validation set is used to compute the lowest significance level so that ICP will not produce sets of multiple predictions. The results for the different NCMs are presented in Table \ref{tab:baseline}. We see that even though the computed significance level does not produce any prediction with multiple classes the error rate is higher than the significance level. This happens because many of the signs are of bad quality which leads to empty prediction sets. This observation model is not reliable.

\vspace{0.05in}

\begin{figure*}[tp]
  \subfloat[]{\label{rev}
      \includegraphics[width=.49\textwidth]{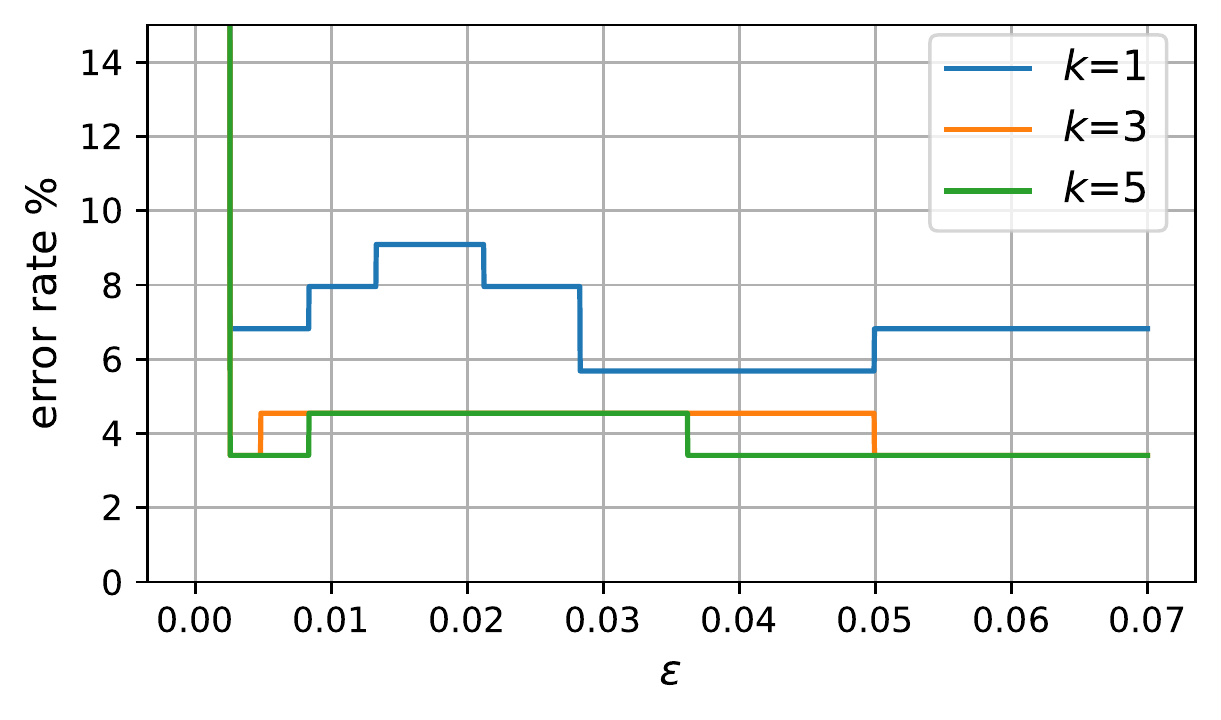}}
    \hfill
  \subfloat[]{\label{rev_sol}
      \includegraphics[width=.49\textwidth]{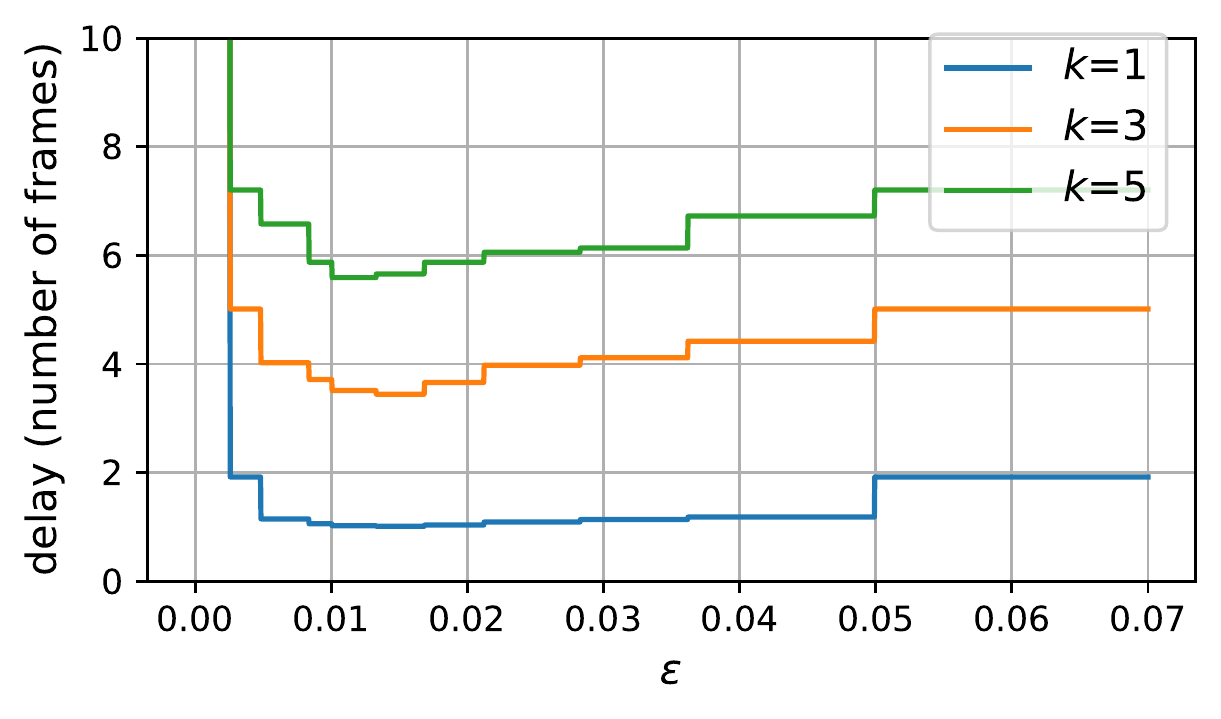}}

  \caption{(a) Error-rate and (b) average number of frames until a decision.} \label{fig:delay}
\end{figure*}

\noindent
\textbf{Improving Prediction Accuracy}
We can improve the LEC classification performance using the feedback loop as described in Section \ref{sec:feedback}. As we can see in Figure~\ref{fig:sequence_error}, the first frames of a sign tend to have more incorrect predictions as they have lower resolution and they lack details. Based on the feedback loop, the LEC uses a new input from the camera until the prediction remains the same for $k$ consecutive frames.
%The choice of $k$ is a trade-off between robustness of the model and classification time. 
Experimenting with different values of $k$, Figure~\ref{fig:delay} shows that as $k$ increases, the error-rate decreases for most of the $\epsilon$ values but the number of frames required to take a decision increases. When $\epsilon<0.003$ the classifier enhanced with the feedback loop could not reach a decision. We also evaluate the efficiency of this classifier regarding to the real-time requirements. A decision for each new sensor query takes on average 1ms, which can be used with typical video frame rates. The memory required to apply the method consists of the memory used to store the representations of the proper training set and the nonconformity scores of the calibration data (45.9MB) and the memory used to store the triplet network (28.5MB) for a total of 74.4MB.  
%when $k>3$ there are no errors or times that a prediction cannot be made in the 43 test sign sequences. As we can see in Figure \ref{fig:delay}. The choice of the significance level is based on the desired classification time.
%\vspace{-6mm}
%\vspace{-6mm}

\section{Conclusions}
\label{sec:conclusion}
The  ICP framework can be used to produce prediction sets that include the correct class with a given confidence. When the inputs to the system are sequential and not IID, applying ICP is not straightforward. Motivated by DDDAS, we design a feedback loop for handling sequential inputs by querying the sensors when a confident prediction cannot be made.  The evaluation results demonstrate that when the inputs to the autonomous system are not IID, the error-rate cannot be bounded. However, the addition of the feedback loop can lower the error-rate by classifying a number of consecutive inputs until a confident decision can be made. The running time and memory requirements indicate that this approach can be used in real-time applications.




% LNCS Homepage, \url{http://www.springer.com/lncs}. Last accessed 4
% \end{thebibliography}

\bibliographystyle{splncs04}
\bibliography{main}

\begin{thebibliography}{10}
\providecommand{\url}[1]{\texttt{#1}}
\providecommand{\urlprefix}{URL }
\providecommand{\doi}[1]{https://doi.org/#1}

\bibitem{ALLAIRE20131959}
Allaire, D., Chambers, J., Cowlagi, R., Kordonowy, D., Lecerf, M., Mainini, L.,
  Ulker, F., Willcox, K.: An offline/online dddas capability for self-aware
  aerospace vehicles. Procedia Computer Science  \textbf{18},  1959 -- 1968
  (2013), 2013 International Conference on Computational Science

\bibitem{Aved2015}
Aved, A., Blasch, E.: Multi-int query language for dddas designs. Procedia
  Computer Science  \textbf{51},  2518--2532 (12 2015)

\bibitem{balasubramanian2014conformal}
Balasubramanian, V., Ho, S.S., Vovk, V.: Conformal Prediction for Reliable
  Machine Learning: Theory, Adaptations and Applications. Morgan Kaufmann
  Publishers Inc., San Francisco, CA, USA, 1st edn. (2014)

\bibitem{10.1117/12.2016338}
Blasch, E., Seetharaman, G., Darema, F.: {Dynamic Data Driven Applications
  Systems (DDDAS) modeling for automatic target recognition}. In: Sadjadi,
  F.A., Mahalanobis, A. (eds.) Automatic Target Recognition XXIII. vol.~8744,
  pp. 165 -- 174. International Society for Optics and Photonics, SPIE (2013)

\bibitem{boursinos2020assurance}
Boursinos, D., Koutsoukos, X.: Assurance monitoring of cyber-physical systems
  with machine learning components. arXiv preprint arXiv:2001.05014  (2020)

\bibitem{boursinos2020trusted}
Boursinos, D., Koutsoukos, X.: Trusted confidence bounds for learning enabled
  cyber-physical systems. arXiv preprint arXiv:2003.05107  (2020)

\bibitem{1398021}
{Darema}, F.: Grid computing and beyond: The context of dynamic data driven
  applications systems. Proceedings of the IEEE  \textbf{93}(3),  692--697
  (2005)

\bibitem{fujimoto2016dynamic}
Fujimoto, R.M., Celik, N., Damgacioglu, H., Hunter, M., Jin, D., Son, Y.J., Xu,
  J.: Dynamic data driven application systems for smart cities and urban
  infrastructures. In: 2016 Winter Simulation Conference (WSC). pp. 1143--1157.
  IEEE (2016)

\bibitem{Guo:2017:CMN:3305381.3305518}
Guo, C., Pleiss, G., Sun, Y., Weinberger, K.Q.: On calibration of modern neural
  networks. In: Proceedings of the 34th International Conference on Machine
  Learning - Volume 70. pp. 1321--1330. ICML'17, JMLR.org (2017)

\bibitem{hoffer2014deep}
Hoffer, E., Ailon, N.: Deep metric learning using triplet network. In:
  International Workshop on Similarity-Based Pattern Recognition. pp. 84--92.
  Springer (2015)

\bibitem{papadopoulos2008inductive}
Papadopoulos, H.: Inductive conformal prediction: Theory and application to
  neural networks. In: Tools in artificial intelligence. IntechOpen (2008)

\bibitem{papernot2018deep}
Papernot, N., McDaniel, P.: Deep k-nearest neighbors: Towards confident,
  interpretable and robust deep learning. arXiv preprint arXiv:1803.04765
  (2018)

\bibitem{Platt99probabilisticoutputs}
Platt, J.C.: Probabilistic outputs for support vector machines and comparisons
  to regularized likelihood methods. In: Advances In Large Margin Classifiers.
  pp. 61--74. MIT Press (1999)

\bibitem{7471414}
{Uzkent}, B., {Hoffman}, M.J., {Vodacek}, A.: Integrating hyperspectral
  likelihoods in a multidimensional assignment algorithm for aerial vehicle
  tracking. IEEE Journal of Selected Topics in Applied Earth Observations and
  Remote Sensing  \textbf{9}(9),  4325--4333 (2016)

\bibitem{xuan2019improved}
Xuan, H., Stylianou, A., Pless, R.: Improved embeddings with easy positive
  triplet mining. arXiv preprint arXiv:1904.04370  (2019)

\bibitem{Zadrozny:2002:TCS:775047.775151}
Zadrozny, B., Elkan, C.: Transforming classifier scores into accurate
  multiclass probability estimates. In: Proceedings of the Eighth ACM SIGKDD
  International Conference on Knowledge Discovery and Data Mining. pp.
  694--699. KDD '02, ACM, New York, NY, USA (2002)

\end{thebibliography}
\end{document}